\documentclass[a4paper, 10pt, conference]{ieeeconf}      
\usepackage{FG2019}

\FGfinalcopy 

\IEEEoverridecommandlockouts                              
\usepackage{graphics} 
\usepackage{epsfig} 
\usepackage{mathptmx} 
\usepackage{times} 
\usepackage{amsmath} 
\usepackage{amssymb}  
\usepackage{subfigure}
\usepackage{graphicx}
\usepackage{adjustbox}
\usepackage{hhline}
\usepackage{tabularx}
\usepackage{verbatim}

\title{\LARGE \bf
Aff-Wild2: Extending the Aff-Wild Database for Affect Recognition
}

\begin{document}

\ifFGfinal

\author{\parbox{16cm}{\centering
    {\large Dimitrios Kollias$^1$ and Stefanos Zafeiriou$^{1,2}$}\\ 
    {\normalsize
    $^1$  Department of Computing, Imperial College London, UK \\
    $^2$ Centre for Machine Vision and Signal Analysis, University of Oulu, Finland}
    }
}
\thispagestyle{empty}
\pagestyle{empty}
\fi
\maketitle

\begin{abstract}
Automatic understanding of human affect using visual signals is a problem that has attracted significant interest over the past 20 years.
However, human emotional states are quite complex. To appraise such states displayed in real-world settings, we need expressive emotional descriptors that are capable of capturing and describing this complexity. The circumplex model of affect, which is described in terms of valence (i.e., how positive or negative is an emotion) and arousal (i.e., power of the activation of the emotion), can be used for this purpose.
Recent progress in the emotion recognition domain has been achieved through the development of deep neural architectures and the availability of very large training databases. To this end, Aff-Wild has been the first large-scale "in-the-wild" database, containing around 1,200,000 frames. In this paper, we build upon this database, extending it with 260 more subjects and 1,413,000 new video frames. We call the union of Aff-Wild with the additional data, Aff-Wild2. The videos are downloaded from Youtube and have large variations in pose, age, illumination conditions, ethnicity and profession. Both  database-specific  as  well  as  cross-database experiments are performed in this paper, by utilizing the Aff-Wild2, along with the RECOLA database. The developed deep neural architectures are based on the joint training of state-of-the-art convolutional and recurrent neural networks with attention mechanism; thus exploiting both the invariant properties of  convolutional  features,  while   modeling  temporal dynamics that arise in human behaviour via the recurrent layers. The obtained results show premise for utilization of the extended Aff-Wild, as well as of the developed deep neural architectures for visual analysis of human behaviour in terms of continuous emotion dimensions.
\end{abstract}

\section{INTRODUCTION}

The field of Automatic Facial Expression Analysis has grown rapidly in recent years, with applications spread across a variety of fields, such as medicine \cite{tagaris1,tagaris2}, health\cite{kollias13}, monitoring, entertainment, lie detection \cite{zhou2015lie,kollias12}. There are two major emotion computing models according to
theories in psychology research \cite{marsella2014computationally}: discrete and dimensional theory. 

Most of the past research has revolved around the recognition of the so-called six universal expressions \cite{dalgleish2000handbook,cowie2003describing,kollias6,kollias10,kollias11}, which are intuitive and simple, but cannot express complex affective states. Also, some Facial Expression Recognition and Analysis systems proposed in the literature focused on the binary occurrence of expressions as FACS Action Units (AUs) \cite{ekman2002facial}, assuming that the expression intensity is fixed. Some representative datasets developed in labs, which are still used in many recent works \cite{jung2015joint}, are the Cohn-Kanade database \cite{tian2001recognizing,lucey2010extended}, the MMI database  \cite{pantic2005web,valstar2010induced}, the Multi-PIE database \cite{gross2010multi} and the BU-3D/BU-4D databases \cite{yin20063d,yin2008high}.
Most lab-controlled databases focused on detecting the occurrence of expressions, regardless of the significant differences in appearance, shape, and temporal dynamics caused by different expression intensities. In reality, expressions can vary greatly in intensity, which constitutes a crucial cue for the interpretation of expressions. 

\begin{figure}[h]
\centering
\adjincludegraphics[height=5.5cm,width=5.5cm]{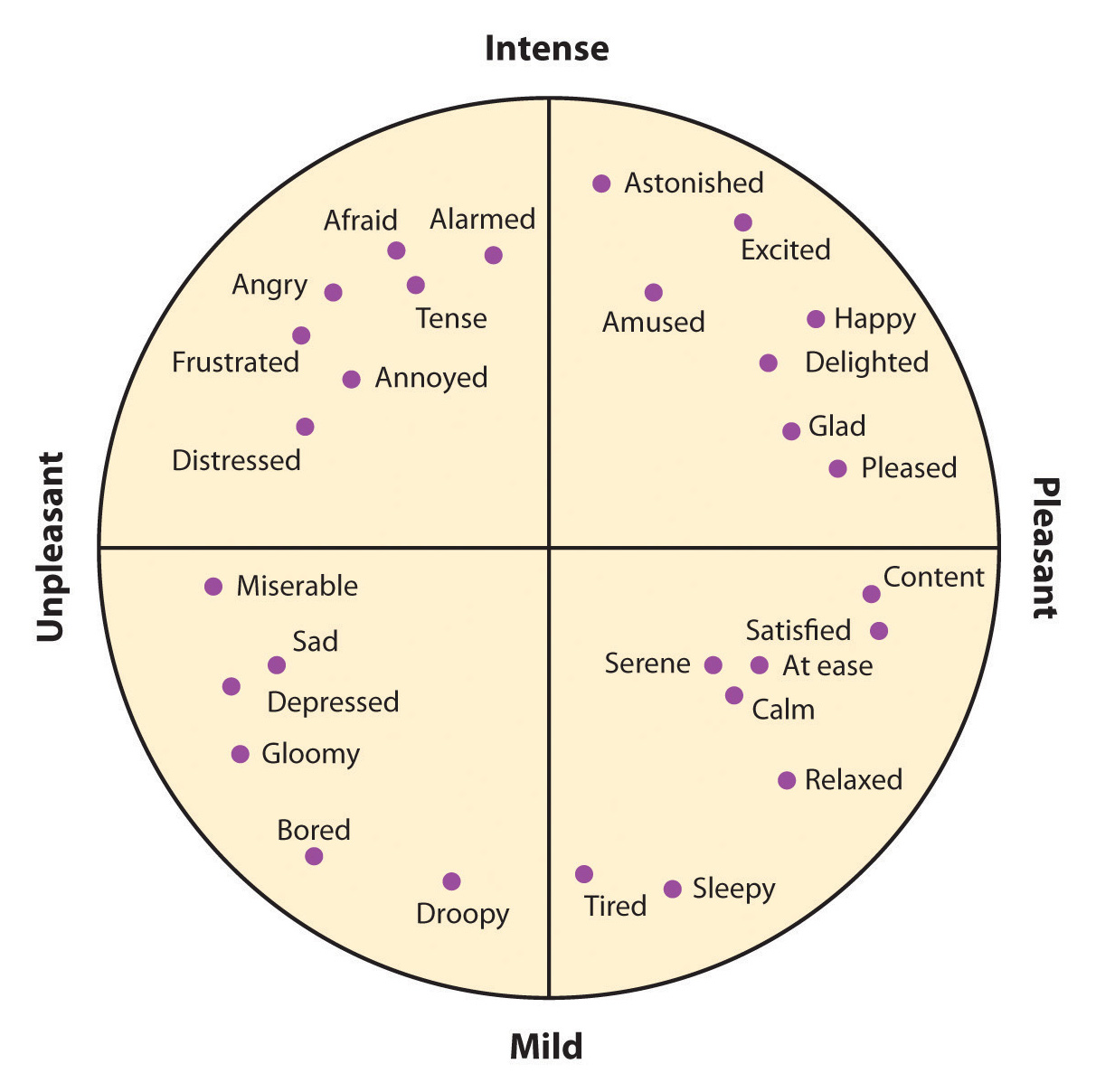}
\caption{The 2D Emotion Wheel, or in other words, the 2D Valence-Arousal Space}
\label{2d-wheel}
\end{figure}

That is why, more recently, affective computing researches are shifting towards dimensional emotion analysis for better
understanding of human emotions. Dimensional theory \cite{whissel1989dictionary},\cite{russell1978evidence} considers an emotional state as a point in a continuous space, the 2D Emotion Wheel \cite{plutchik1980emotion}, shown in Figure \ref{2d-wheel}.  The  dimensions represent arousal, indicating the level of affective activation, and valence, measuring the level of pleasure. Hence, dimensional theory can model subtle,
complicated, and continuous affective behaviors.

Various emotion recognition databases have been developed utilizing the dimensional emotion representation. 
Examples are the SAL \cite{douglas2008sensitive}, SEMAINE \cite{mckeown2012semaine}, MAHNOB-HCI \cite{soleymani2012multimodal}, Belfast naturalistic \footnote{\label{note1}https://belfast-naturalistic-db.sspnet.eu/}, Belfast induced \cite{sneddon2012belfast}, DEAP \cite{koelstra2012deap}, RECOLA \cite{ringeval2013introducing}, SEWA \footnote{\label{note2}http://sewaproject.eu} databases. Nevertheless, the majority of collected data this far, although containing naturalistic emotional states, have been captured in highly controlled recording conditions and have rather small size. Moreover, they display very low diversity in terms of the total number of subjects they contain; they include a limited amount of head pose variations and occlusions, static backgrounds and uniform illuminations. Recently, Aff-Wild was created \cite{kollias1,kollias3}, constituting the 
first large-scale "in-the-wild" database, with over 60 hours of video data, annotated in terms of the valence-arousal dimensions. 

\begin{figure*}
\centering
\scalebox{0.9}{
\begin{tabular}{c}
  \includegraphics[height=1.5cm]{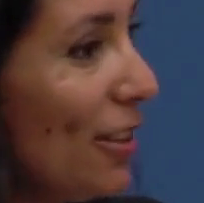}\includegraphics[height=1.5cm]{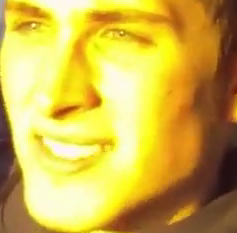}\includegraphics[height=1.5cm]{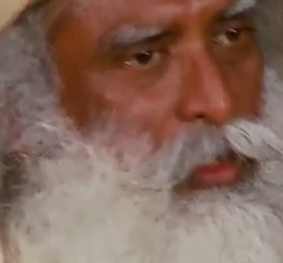}\includegraphics[height=1.5cm]{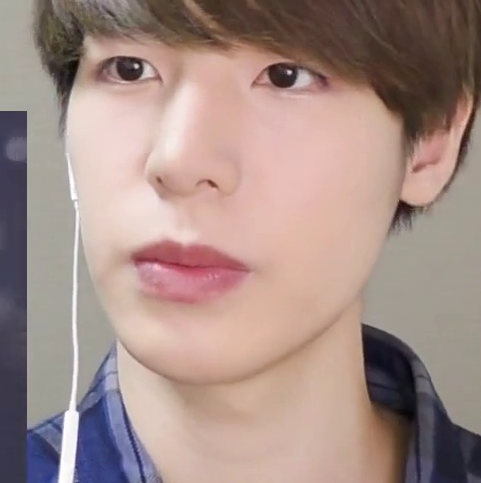}\includegraphics[height=1.5cm]{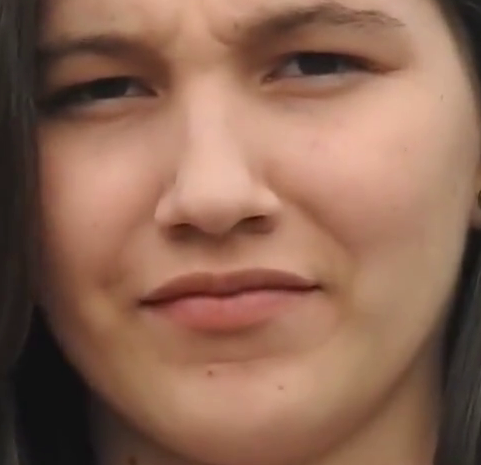}\includegraphics[height=1.5cm]{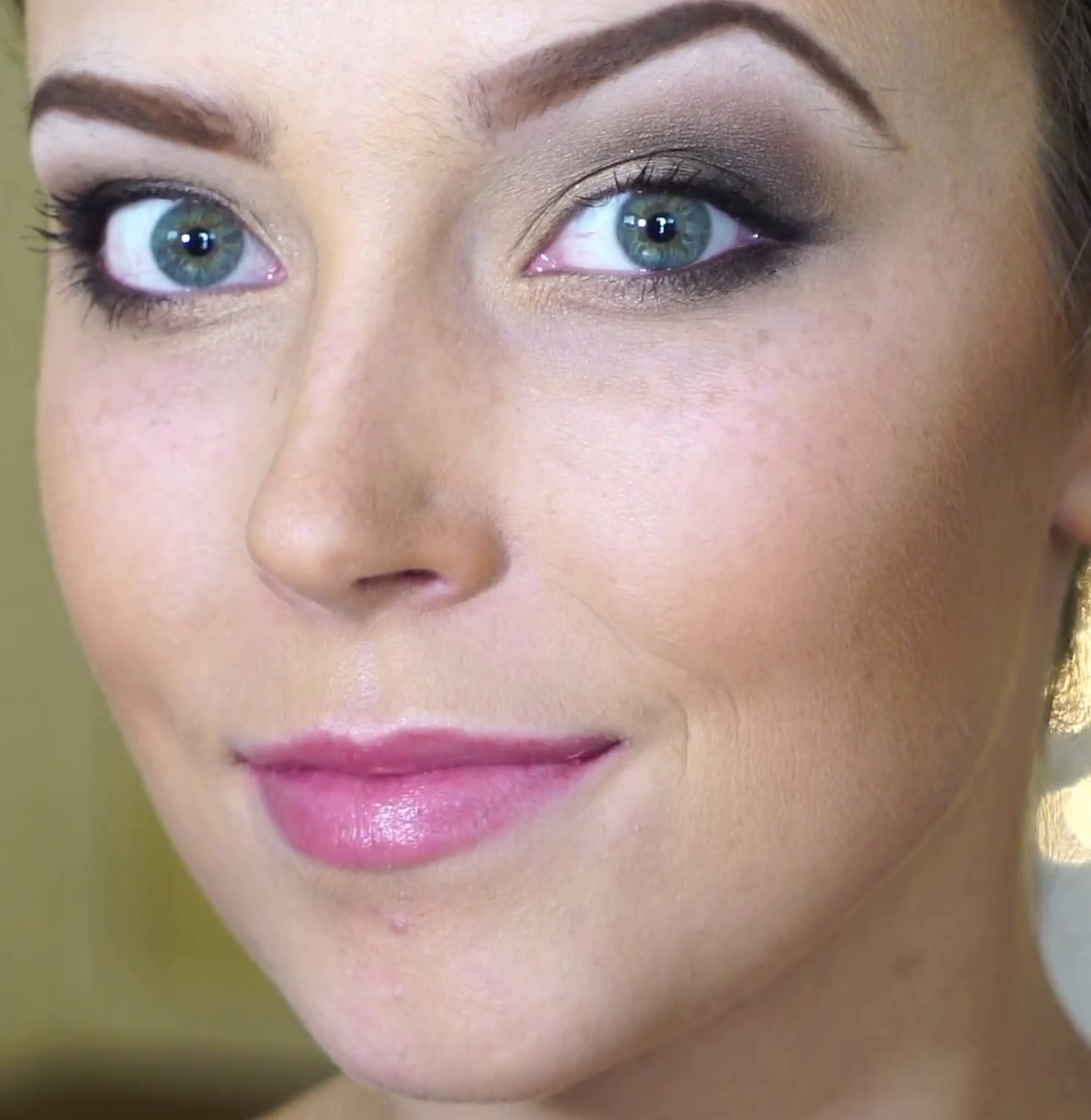}\includegraphics[height=1.5cm]{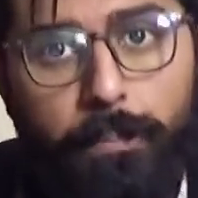}\includegraphics[height=1.5cm]{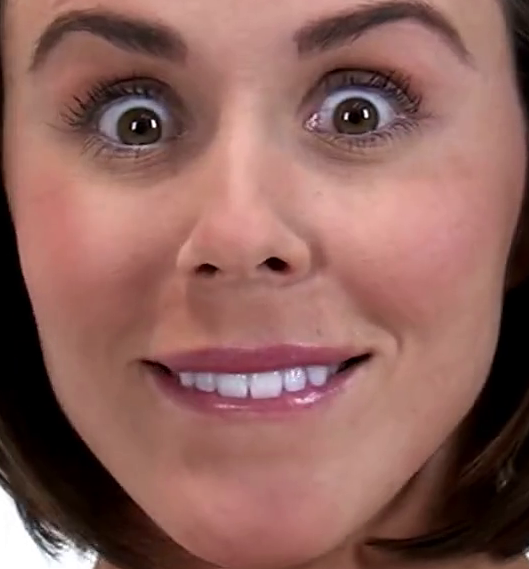}\includegraphics[height=1.5cm]{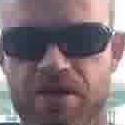}\includegraphics[height=1.5cm]{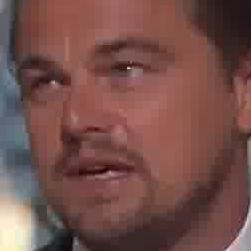}\\
\includegraphics[height=1.5cm]{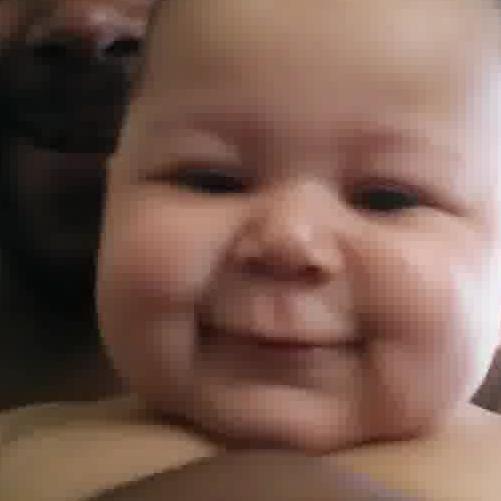}\includegraphics[height=1.5cm]{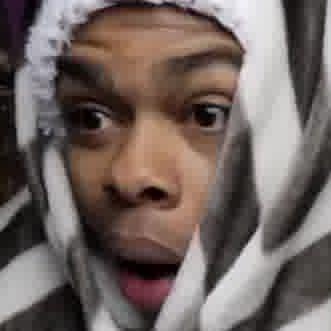}\includegraphics[height=1.5cm]{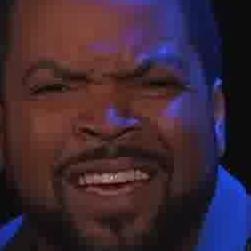}\includegraphics[height=1.5cm]{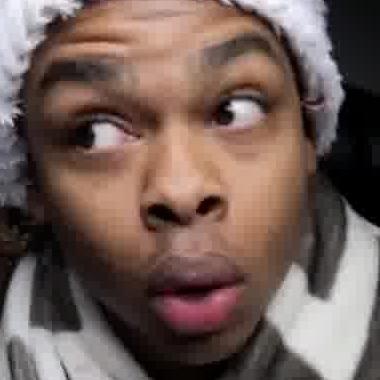}\includegraphics[height=1.5cm]{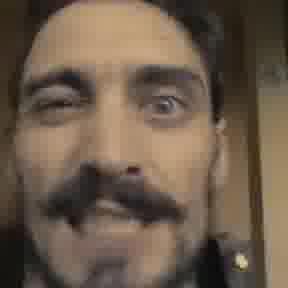}\includegraphics[height=1.5cm]{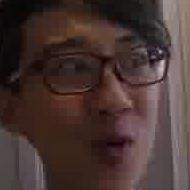}\includegraphics[height=1.5cm]{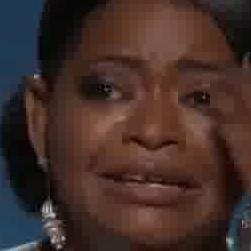}\includegraphics[height=1.5cm]{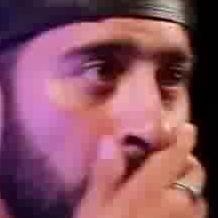}\includegraphics[height=1.5cm]{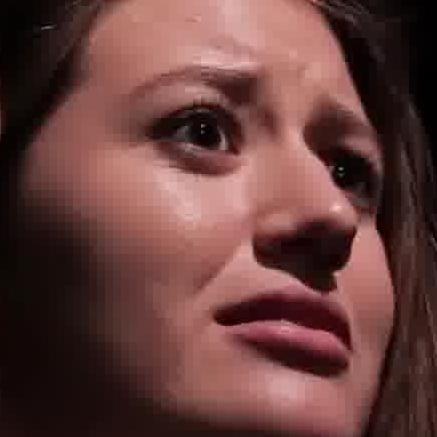}\includegraphics[height=1.5cm]{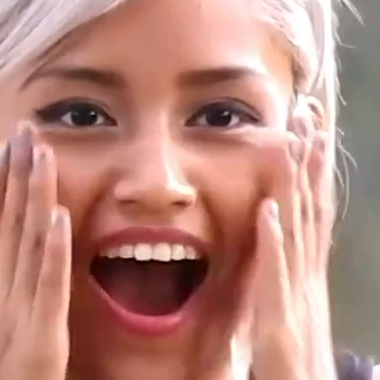} \\
\end{tabular}
}
\caption{Some frames from the additional data that formed Aff-Wild2, showing subjects of different ethnicities and age groups, in different emotional states, in a variety of head poses, illumination conditions and occlusions.}
\label{frames_from_db}
\end{figure*}

To this end, in this paper, our contribution is a significant extension of the Aff-Wild database, by approximately doubling the
number of included video frames and the number of subjects; thus, improving the variability of the included behaviors and of the involved persons. We call the union of the existing Aff-Wild with the additional data, Aff-Wild2. Let us note here that this constitutes a preliminary work of \cite{kollias15,kollias5}. 
We then present end-to-end deep neural architectures trained over Aff-Wild2 and used for estimation of the valence and arousal emotion dimensions. The proposed deep architectures are composed of convolutional and recurrent neural network (CNN-RNN) layers, along with an attention mechanism.  

Another contribution of the paper is that by training deep neural networks on the Aff-Wild2 and by fine-tuning them on another existing databases, such as RECOLA, we achieve 
state-of-the-art emotion recognition performance on the other database, substantially exceeding the performance of other networks trained with that database.

All of the obtained results illustrate the very good performance of the presented approach and its potential for effective visual analysis of human behavior in terms of continuous emotion dimensions.

Section \ref{two} of the paper presents the Aff-Wild database and then continues with a detailed description of the additional dataset that was collected, illustrating the related attributes, the pre-processing and annotation steps. Section \ref{three} describes the facial and emotion analysis, focusing on the deep neural architectures and the evaluation metrics that were used. Section \ref{four} presents the experimental study and the obtained results. Conclusions and further work are described in Section \ref{five} of the paper.




\section{The Aff-Wild2 database}\label{two}

\subsection{The existing Aff-Wild database}

The Aff-Wild has been the first large scale captured in-the-wild database, consisting of 298 videos and displaying reactions
of 200 subjects, with a total video duration of more than 30 hours. The total number of frames in this database is 1,180,000. Regarding subjects' gender, 130 of them are male and 70 female.
This database has been annotated by 8 lay experts with regards to valence and arousal. The Aff-Wild database served as benchmark for the Aff-Wild Challenge, organized in conjunction
with CVPR 2017. The aim for this database was to collect spontaneous facial behaviors in arbitrary recording conditions. To
this end, the videos were collected using the Youtube video
sharing web-site. The main keyword that was used to retrieve
the videos was ”reaction”.

\subsection{Additional dataset and properties}
We collected a new dataset consisting of $260$ videos, having $1,413,000$ 
frames and a total length of $13$ hours and $5$ minutes. 
$11$ out of those 260 videos displayed two subjects, all of which have been annotated. In total, the new dataset shows 258 
subjects, 
 with 149  being male and 109 female. 
Table \ref{attrs} shows the general attributes of the additional data added to the Aff-Wild, forming the Aff-Wild2.

The aim of adding these new videos in the existing Aff-Wild has been to extend the spontaneous facial behaviors in arbitrary recording conditions met in Aff-Wild, also significantly increasing the number of different subjects in it. All these additional videos have wide range in subjects': age (from babies to young children to elderly people); ethnicity (subjects are caucasian, hispanic or latino, asian, black, or african american); profession (e.g. actors, athletes, politicians, journalists); head pose; illumination conditions; occlussions; emotions.

\begin{table}[h]
\caption{General Attributes of the additional data added to the Aff-Wild, forming Aff-Wild2}
\label{attrs}
\centering
\begin{tabular}{|c|c|}
\hline
Attribute & Description \\ 
\hhline{|=|=|}
No of frames  & $1,413,000$ \\
\hline
No of videos & $260$ \\
\hline
No of subjects  & $258$ ($149$ male; $109$ female)  \\
\hline
No of annotators  & $4$ \\
\hline
Length of videos & $3$ secs $-$ $15$ mins $4$ secs  \\
\hline
Video format & MP$4$\\
\hline
Mean Image Resolution & $1454 \times 890 $\\
\hline
\end{tabular}
\end{table}

Figure \ref{frames_from_db} shows some frames from the additional data, verifying the in-the-wild nature of this set, in which people of different ethnicities and age groups display various emotions, with different head poses and facial occlusions and under different illumination conditions.

The videos were collected using the Youtube video sharing web-site. All of the collected videos were provided under Creative Commons licence. The keywords that were used to retrieve the videos were mainly "reaction" and other emotion related words from the 2-D Emotion Wheel, shown in Figure \ref{2d-wheel}.

The videos show subjects who: react on positive and unexpected surprise (i.e., receiving gifts); are stand-up comedians; give a really interesting speech in ceremonies (i.e. oscar awards, united nations); participate in interviews (i.e., job or football or news ones); are taking an oral exam; react to flirt, or rejection; are in a car that runs really fast; are (movie) stars and react to mean, or funny tweets; are in a talk show; are giving an audition for a role; are giving lectures on depression, or other serious disorders; react on something that brings them happiness, or fulfillment (i.e., seeing their newborn baby for the first time); react on important political issues; are performing activities that are either passive, boring, apathetic, or intense (i.e., yoga, anxiety and anger control, roller-coaster); are watching a highly anticipated trailer, or horror movie; are teaching a new language.

All the videos are in MP4 format and have been annotated in terms of valence and arousal.
Four subjects have annotated the videos using the method proposed in \cite{cowie2000feeltrace}. That is, an on line annotation procedure was used, according to which annotators were watching each video and provided their annotations through a joystick. Valence and arousal values ranged continuously in [$-1 $, $+1 $].

\begin{figure}[h]
\centering
\scalebox{0.92}{
\adjincludegraphics[height=6cm,width=9.4cm]{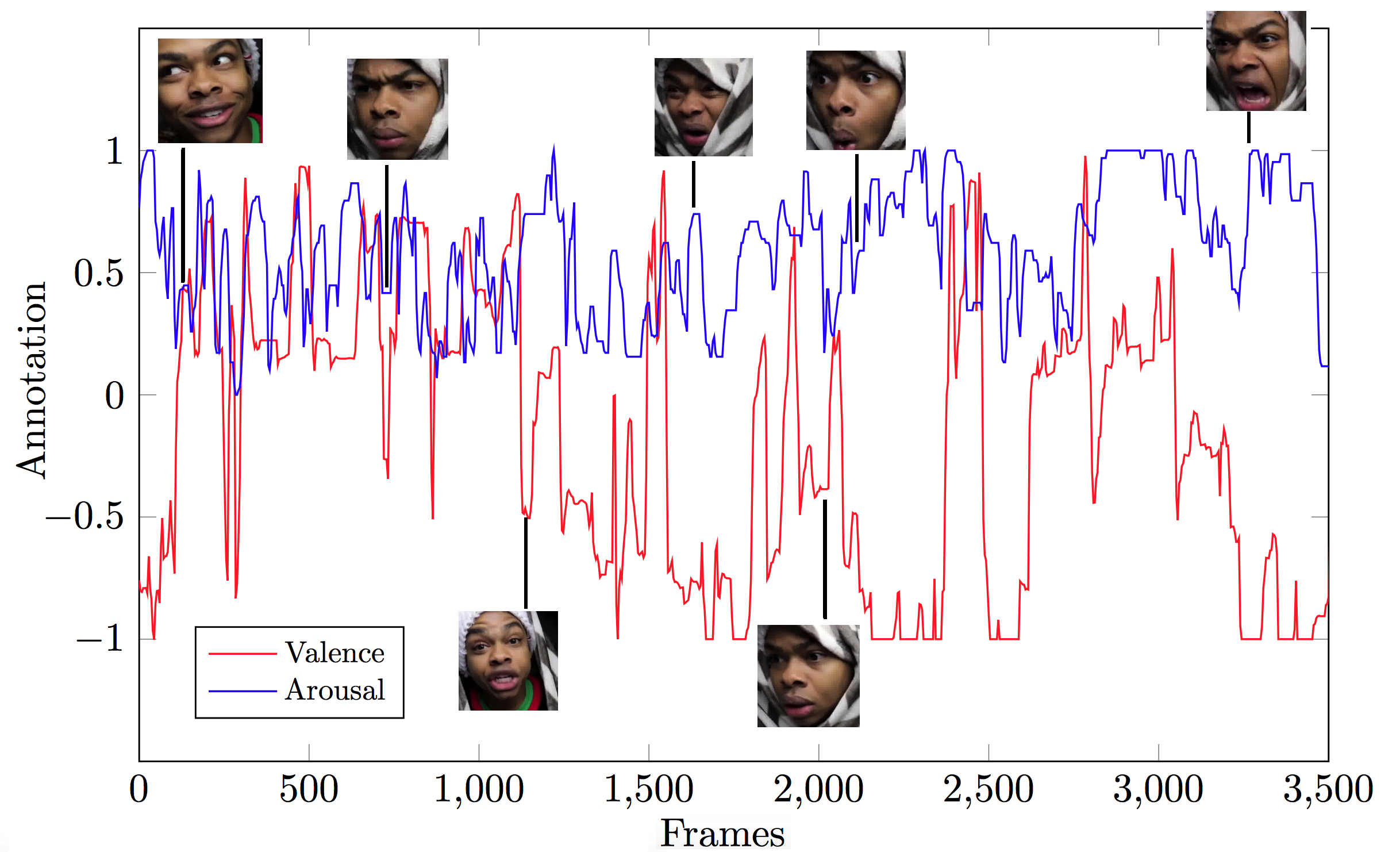}
}
\caption{Valence and arousal annotations over a part of a video, along with corresponding frames, illustrating the in-the-wild
nature of the additional dataset (different emotional states, rapid emotional changes, occlusions)}
\label{trajectory}
\end{figure}

Figure \ref{trajectory} shows an example of annotated valence and
arousal values over a part of a video in the additional data, together
with some corresponding frames. This illustrates the in-the-wild
nature of the database, namely, including many different
emotional states, rapid emotional changes and occlusions
in the facial areas. 

\begin{figure}
\centering
\begin{tabular}{cc}
\adjincludegraphics[height=4.5cm,width=6.5cm]{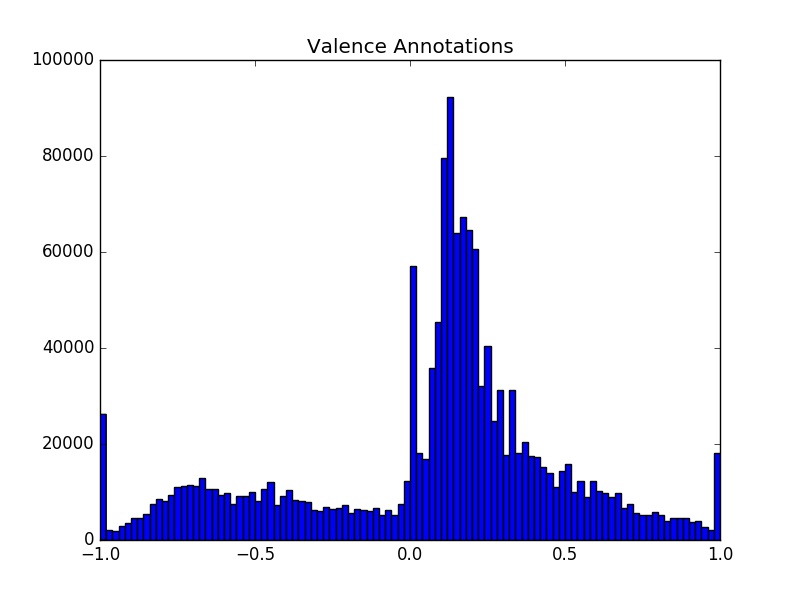}\\
\adjincludegraphics[height=4.5cm,width=6.5cm]{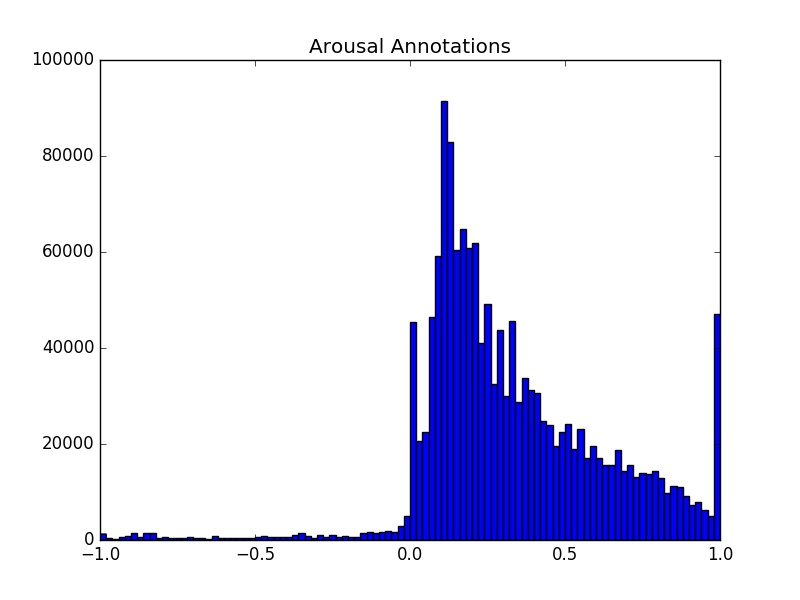}\\
\end{tabular}
\caption{Histogram of valence and arousal annotations of the additional data added in Aff-Wild database and shaping Aff-Wild2.}
\label{hist}
\end{figure}

Figure \ref{hist} provides a histogram of the annotated
values for valence and arousal in the generated additional dataset. Since we have focused on users' reactions, arousal is generally positive; positive reactions were the majority, resulting in more positive than negative valence values.

\subsection{Data pre-processing}

\subsubsection{Pre-processing at video level}

At first, 241 videos have been downloaded from Youtube selecting the best video (image and audio) quality. These videos were then converted into MP4 format using FormatFactory. Using VirtualDub \cite{lee2002welcome}, the videos were processed and trimmed, especially at the beginning and ending, so as to remove unnecessary content (i.e., advertisements, scenes with no human face, or with only captions). 

After that, some videos were split into smaller videos because: (i) they included individual scenes with different people, or (ii) in terms of the time dimension, they displayed a person into different, non-continuous time instances. An example of case (i) is a video showing the reactions of 7 people inside a car, showing a different scene for each person; this video was separated into 7 videos,  independent from each other. An example of case (ii) is a video consisting of two parts, with the first recorded in day light and the second recorded at night. This video was split in two new videos, since the person appearing in them was not continuously expressing an emotion.  

It is worth mentioning that when two people appear simultaneously in the video (e.g., having a discussion), this video was not split in two new ones, but has been retained as it was. We allowed such cases, so that the emotion recognition system can learn from the interaction between those two persons. In our database, we had 11 such cases and all subjects appearing in the videos have been annotated.
After this step, as already mentioned, the generated dataset included 260 videos in total.

\subsubsection{Frame-rate conversion}

Each of those 260 videos had a different frame per second (fps) value. As can be seen in Table \ref{fps}, the majority of the videos had a fps rate around 30. To have the same frame rate over all videos, we converted them to have the same fps value, equal to 30.

\begin{table}
\caption{Number and duration of videos found in the additional data, along with their corresponding fps rates.}
\label{fps}
\centering
\scalebox{0.88}{
\begin{tabular}{|c|c|c|c|c|c|c|}
\hline
Fps value &30 & 29 & 25 & 60 & 15 & 17\\
                  \hline
No of videos  & 168  & 10 & 60  & 15 & 5 & 2    \\
\hline
Duration          & 8:25:00  & 0:35:00  & 2:10:00  & 1:35:00  & 0:10:00 & 0:05:00 \\
\hline
\end{tabular}
}
\end{table}

\subsection{Data annotation}

\subsubsection{The annotation procedure}

For data annotation, we used the same tool as \cite{zafeiriou2017aff}  that
builds on other existing ones, specifically Feeltrace \cite{cowie2000feeltrace} and Gtrace \cite{cowie2012tracing}. A time-continuous annotation is performed for each affective dimension.

It should be mentioned that the annotation tool also has 
the ability to show the inserted valence and arousal values while displaying a respective video. This was used for
annotation verification, as described below, in the annotation post-processing step.

Four experts were chosen to perform the annotation task. All annotators were computer scientists who had worked on face
analysis problems and all had a working understanding of facial expressions. Each
annotator was instructed orally and through a multi-page
document on the procedure to follow for the task. This document
included a list of some well identified emotional cues
for both arousal and valence, providing a common basis for
the annotation task. On top of that the experts used their own
appraisal of the subjects' emotional states for creating the annotations.
Before starting the annotation of each video, the
experts watched the whole video, so as to know what to expect
regarding the emotions being displayed in the video.

\subsubsection{Annotation post-processing}

After the annotators had completed the insertion of the selected valence and arousal values in each video, a post-processing annotation verification step was performed.

Every expert-annotator watched all videos for a second
time, in order to verify that the recorded annotations
were in accordance with the shown emotions in the videos, 
or decide to change the annotations accordingly. In this way, a further validation of annotations was achieved.
After the annotations have been validated by the annotators,
a final annotation selection step followed.

\begin{figure}[h]
\centering
\adjincludegraphics[height=5.5cm,width=8.5cm]{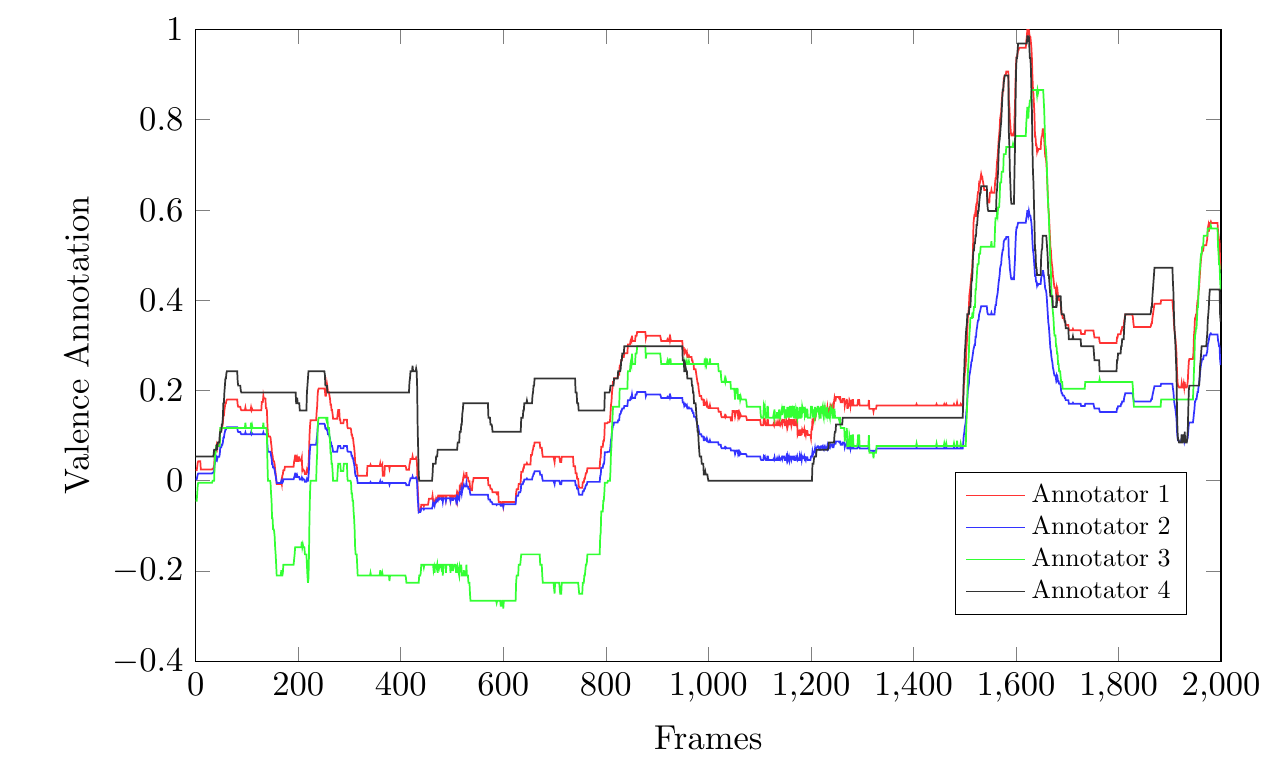}
\caption{All four valence annotations in a video segment. The value of MAIC (mean of average inter annotation correlation) is 0.64 which is similar to the mean MAIC obtained over all additional data.}
\label{annotations}
\end{figure}

Two different approaches were considered for determining the final label values. Those were: (i) computing the mean of the 4 annotators, (ii) performing median filtering for each video, on each expert annotation and then computing the mean.
To select one of them, we performed, independently for valence and arousal the following procedure:
At first, we computed the inter-annotator correlations, i.e., the correlations of each one of the four annotators with all other annotators, over all videos; this  resulted in three correlation values per annotator. Then, we computed for each annotator, his/her average inter-annotator correlations, resulting in one pair of (valence, arousal) values per annotator. After that, we computed the mean of those pairs of values; let us denote the resulting mean as MAIC, meaning mean of average inter annotation correlation.
In approach (i), MAIC was (0.60, 0.58) for valence-arousal. In approach (ii), we experimented with different values of the median filter size, in particular 5, 15 and 30 frames (0.1, 0.5 and 1 sec, respectively). We did not examine higher values, to avoid severely changing the annotators' values. The corresponding MIAC values for valence-arousal were (0.603, 0.582), (0.62, 0.594) and (0.63, 0.60). We selected the latter approach, since it provided a slightly better MAIC value. An example set of annotations is shown in Figure \ref{annotations}, in an effort to clarify the obtained MAIC values. It shows the four annotations in a video segment for valence, with MAIC value of 0.64 (similar to the value obtained over all additional data).

\subsection{The properties of Aff-Wild2 database}

Since all additional videos are in MP4 format and have a frame rate of 30, the videos of the original Aff-Wild database have also been converted to MP4 format with a fps rate equal to 30. Next, we concatenated the additional dataset with the Aff-Wild database, creating a new database, which we call Aff-Wild2.
In total, Aff-Wild2 consists of $558$ videos with $2,786,201$ 
frames. The total number of subjects is 458, with 279 of them being male and 179 being female. 

\subsection{Database partition sets}\label{sets}

The Aff-Wild2 database was also split into three sets: training, validation and test sets. The partitioning was done in a subject independent manner, in the sense that a  person could appear only in one of those three sets and not in all, or in any combination of them. The resulting training, validation and test sets consisted of 350, 70, 138 videos, with $1,601,000$ , $405,000$ and $780,201$ frames, respectively.

\subsection{Other Databases: RECOLA}

The REmote COLlaborative and Affective (RECOLA) dataset is a corpus that monitors subjects collaborating in dyadic teams. This corpus includes multimodal data, i.e. audio, video, ECG and EDA (physiological data). 
It consists of 46 French speaking subjects being recorded for 9.5\,h recordings in total. The recordings  have been annotated, in terms of valence and arousal, for 5\,minutes each by 6 French-speaking annotators (three male, three female). The dataset is divided into three parts, namely, training (16 subjects), validation (15 subjects) and test (15 subjects), in such a way that the gender, age and mother tongue are balanced.
Valence and arousal range continuously in [$-1 $, $+1 $].

\section{FACIAL AND EMOTION ANALYSIS} \label{three}
\subsection{Face detection}

The analysis of Aff-Wild2 database starts with the detection of faces \cite{avrithis2000broadcast} in all video frames. We followed the same methodology described in \cite{mathias2014face}, which was also performed in \cite{zafeiriou2017aff} for generating facial bounding boxes in Aff-Wild database.
The resulting facial images were resized to resolution of  $96 \times 96 \times 3$, with their
intensity values being normalized to the range [$-1 $, $+1 $].

\subsection{The CNN-RNN Deep Neural Architectures}\label{diff_architectures}

Since the Aff-Wild2 database consists of videos, we developed CNN-RNN architectures that can exploit the temporal dynamics of the data. The following strategies have been adopted for training all CNN-RNN networks: (i) keep the CNN weights fixed to their pre-trained values and train only the RNN layers; (ii) randomly initialize the RNN part and train the whole CNN-RNN; (iii) initialize the RNN part with the best performing model of (i) and then train the whole CNN-RNN.
In the following, we present details about the CNN and RNN parts of the developed architectures:

\subsubsection{CNN part of the architectures}

For CNN models, we experimented with the VGGFACE\cite{parkhi2015deep}, ResNet-50\cite{he2016deep} and DenseNet-121\cite{iandola2014densenet} structures, which can provide rich representations for the visual emotion analysis task. In the VGGFACE and ResNet-50 cases, we considered three different setting, where networks are: a) pre-trained on the VGGFACE database\cite{parkhi2015deep}, b) pre-trained on the VGGFACE2 database\cite{cao2018vggface2}, and c) pre-trained on the VGGFACE and then on the VGGFACE2 datasets. In the DenseNet-121 case we considered a setting in which the network is pre-trained on the ImageNet database. Those network settings are shown in Table \ref{tab:CNNmodels}, together with the denoting used for referencing these networks in the experimental study.

\begin{table}[h]
\caption{CNN architectures with corresponding pre-trained model}
\centering
\begin{tabular}{|c||c||c|}
\hline{}
Network         & Database pre-trained on & Denote \\ \hline
\begin{tabular}{@{}c@{}} VGGFace \\ or \\  ResNet-50  \end{tabular}  &  \begin{tabular}{@{}c@{}} VGGFace \\ VGGFace2 \\ VGGFace + VGGFace2 \end{tabular} & \begin{tabular}{@{}c@{}} VGGFace1 / ResNet1 \\ VGGFace2 / ResNet2  \\ VGGFace12 / ResNet12   \end{tabular}         \\ \hline
DenseNet-121    & ImageNet   & DenseNet         \\ \hline
\end{tabular}
\label{tab:CNNmodels}
\end{table}

\subsubsection{RNN part of the architectures}

For RNN models, we experimented with Long Short Term Memories (LSTM), Gated Recurrent Units (GRU) and independent RNNs (indRNN). 
We also tested using an attention layer on top of the developed CNN-RNN models.
In the VGGFACE case, the RNN part was fed with the output of either the first fully connected (FC) layer or the last pooling layer after global average pooling.
In all other CNN-RNN architectures, the RNN part was fed with the output of the last convolution or pooling layer of the respective CNN part.
We used 2 hidden layers in the RNN, each having 128 hidden units, as in \cite{kollias7,kollias14}.

\subsubsection{Best performing architecture/model} 
\begin{figure}[h]
\centering
\adjincludegraphics[height=6cm,width=8.cm]{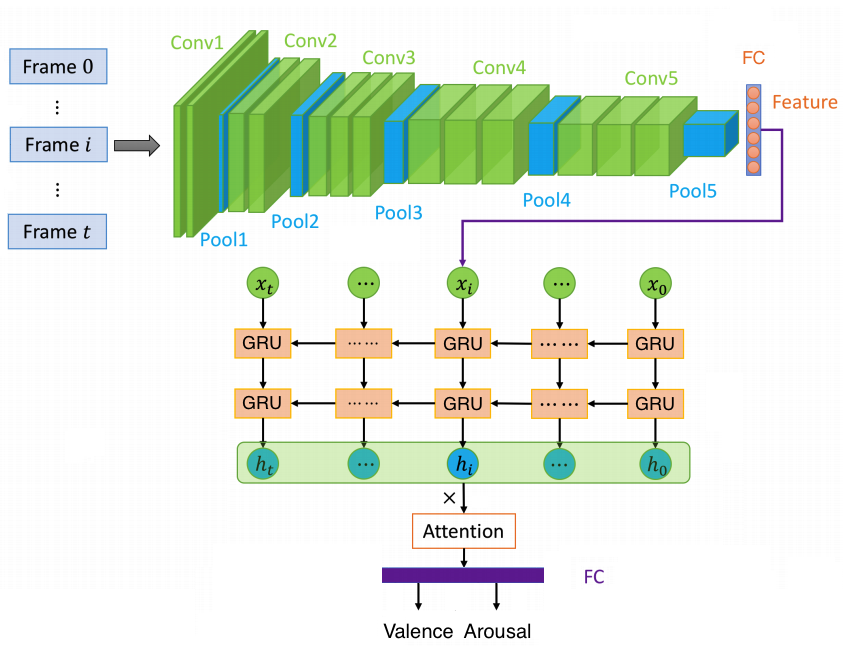}
\caption{The architecture/model that achieved the best results in Aff-Wild2}
\label{architecture}
\end{figure}

Figure \ref{architecture} shows the architecture/model that achieved the best results in Aff-Wild2, as shown in Section \ref{four}. The CNN part is based on the VGGFACE, the RNN part is based on GRUs taking as input the output of the first FC layer of VGGFACE, an attention layer is stacked on top of the RNN and before the output layer that gives the valence and arousal predictions.

\subsection{The evaluation criterion}\label{criteria}

The criterion that is considered for evaluating the performance
of the networks is Concordance Correlation
Coefficient (CCC), which is used in all related Challenges (AVEC \cite{ringeval2017avec}, Affect-in-the-Wild\footnote{https://ibug.doc.ic.ac.uk/resources/first-affect-wild-challenge/}). CCC takes values in the range [$-1$,$+1$], with $+1$
indicating perfect concordance and $-1$ perfect discordance.
The highest the value of the CCC the better the
fit between annotations and predictions, and therefore high
values are desired. CCC is defined as follows:

\begin{equation} \label{eq:1}
\rho_c = \frac{2 s_{xy}}{s_x^2 + s_y^2 + (\bar{x} - \bar{y})^2}
\end{equation}

\noindent
where $s_x$ and $s_y$ are the variances of the ground truth and predicted values respectively, $\bar{x}$ and $\bar{y}$ are the corresponding mean values and $s_{xy}$ is the respective covariance value.

The loss function used for training all the networks was: 

\begin{equation} \label{eq:3}
\mathcal{L}_{total} = 1 - \frac{\rho_a + \rho_v}{2},
\end{equation}

\noindent
where $\rho_a$ and $\rho_v$ are the CCC (defined in eq.(1) ) for the arousal and valence, respectively.

\section{Experimental study} \label{four}

\subsection{Training implementation details}

In order to train all deep neural architectures, we utilized the Adam optimizer
algorithm and used exponential decay for the learning rate in fixed number of
epochs; the batch size was set to to 320 (consisting of 4 different sequences, each one having 80 consecutive frames), the attention length was chosen to be 32, the initial learning rate was set to 0.001, the decay steps for the learning rate were 400 and its decay factor was 0.97.
The platform used for this implementation was Tensorflow.

\subsection{Experiments on Aff-Wild2}

In the experimental study, we evaluate the performance of all deep neural architectures described in Section \ref{diff_architectures}. 
Table \ref{vggface} shows the CCC performance for estimation of valence and arousal values, provided by the above-mentioned CNN-RNN networks in which: a) the CNN part is based on the VGGFACE model and is pre-trained on either the VGGFACE,  or the VGGFACE2,  or first the VGGFACE and then the VGGFACE2 databases; b) the RNN part uses either LSTM, or GRU, or indRNN neuron models. 

It should be mentioned that the case in which the output of the FC layer of VGGFACE is fed to the RNN had better performance compared to the case where the RNN was fed with the output of the last pooling layer after global average pooling. We do not present the performance of the latter case in order not to clutter the results. It can be seen that the VGGFACE1-GRU network (pretrained on VGGFACE database) provided the best performance.

\begin{table}[h]
\caption{Obtained CCC values for valence and arousal for CNN-RNN architectures that are based on the VGGFACE and under different settings described in Section \ref{diff_architectures}. In parenthesis are the performances obtained on the validation set.}
\label{vggface}
\centering
\begin{tabular}{ |c||c|c|  }
 \hline
 \multicolumn{1}{|c||}{Networks} & \multicolumn{2}{c|}{CCC}  \\
 \hline
   & Valence & Arousal  \\
 \hline
VGGFACE1-LSTM & 0.44 (0.49) &  0.34 (0.37)    \\
 \hline
\textbf{VGGFACE1-GRU} &\textbf{  0.52 (0.55)} &\textbf{  0.41 (0.45)}  \\
 \hline
VGGFACE1-indRNN &  0.48 (0.52) &   0.35 (0.39)   \\
 \hline
VGGFACE2-LSTM &0.42  (0.47) &0.33  (0.36)       \\
 \hline
VGGFACE2-GRU &0.46  (0.51) &0.37  (0.40)  \\
 \hline
VGGFACE2-indRNN& 0.44 (0.48) & 0.34   (0.37)    \\
 \hline
VGGFACE12-LSTM & 0.44 (0.48) & 0.33  (0.37)     \\
 \hline
VGGFACE12-GRU &0.47  (0.53) &0.35  (0.42)  \\
 \hline
VGGFACE12-indRNN &0.46  (0.51) &0.35  (0.39)   \\
 \hline
\end{tabular}
\end{table}

Similar results are shown in Table \ref{resnet}. The difference, here, is that the CNN part of the CNN-RNN architecture is based on the ResNet-50 model. It can be seen that the ResNet2-GRU network (pretrained on VGGFACE2 database) provided the best performance.

\begin{table}[h]
\caption{Obtained CCC values for valence and arousal for CNN-RNN architectures that are based on the ResNet and under different settings described in Section \ref{diff_architectures}. In parenthesis are the performances obtained on the validation set.}
\label{resnet}
\centering
\begin{tabular}{ |c||c|c|  }
 \hline
 \multicolumn{1}{|c||}{Networks} & \multicolumn{2}{c|}{CCC}  \\
 \hline
   & Valence & Arousal  \\
 \hline
ResNet1-LSTM & 0.44 (0.47) & 0.39 (0.41)    \\
 \hline
ResNet1-GRU &0.45 (0.49)  &0.39  (0.39)  \\
 \hline
ResNet1-indRNN &0.42 (0.46)  &0.38  (0.39)   \\
 \hline
ResNet2-LSTM & 0.46  (0.49) & 0.38  (0.42)     \\
 \hline
\textbf{ResNet2-GRU} &\textbf{0.50  (0.53)} &\textbf{0.42  (0.44)}  \\
 \hline
ResNet2-indRNN &0.46 (0.50)  &0.39  (0.41)   \\
 \hline
ResNet12-LSTM & 0.44  (0.48) & 0.39  (0.41)     \\
 \hline
ResNet12-GRU &0.48  (0.51)  &0.40  (0.43)  \\
 \hline
ResNet12-indRNN &0.47  (0.49)  &0.36  (0.39)   \\
 \hline
\end{tabular}
\end{table}

The use of DenseNet-121 in the CNN part of the above-examined CNN-RNN architectures was also investigated. Table \ref{densenet} compares the performance between models that their CNN part is based on the DenseNet-121 and the RNN is based on either LSTM, GRU or indRNN neuron models. It can be seen that the DenseNet-GRU network provided the best performance.

\begin{table}[h]
\caption{Obtained CCC values for valence and arousal for CNN-RNN architectures that are based on the DenseNet. In parenthesis are the performances obtained on the validation set.}
\label{densenet}
\centering
\begin{tabular}{ |c||c|c|  }
 \hline
 \multicolumn{1}{|c||}{Networks} & \multicolumn{2}{c|}{CCC}  \\
 \hline
   & Valence & Arousal  \\
 \hline
DenseNet-LSTM & 0.43  (0.48) & 0.38  (0.41)     \\
 \hline
\textbf{DenseNet-GRU} &\textbf{0.48  (0.51)} &\textbf{0.39  (0.43)}   \\
 \hline
DenseNet-indRNN &0.43  (0.47) &0.37   (0.41)  \\
 \hline
\end{tabular}
\end{table}

We then examined the improvement provided by adding an attention mechanism on top of the best performing networks in the above three Tables. Table \ref{all} compares the performance of the best performing VGGFACE1-GRU, ResNet2-GRU and DenseNet-GRU networks with the same networks when they have an attention mechanism on top of their GRU part. 
It can be easily seen that the VGGFACE1-GRU-attention network (that was pre-trained on VGGFACE database and included an attention layer) had the best performance on the test set, providing a 0.55 value for valence and 0.45 for arousal.

\begin{table}[h]
\caption{Obtained CCC values between the best performing CNN-RNN architectures' configurations and same architectures with an attention layer on top of their RNN part. In parenthesis are the performances obtained on the validation set.}
\label{all}
\centering
\begin{tabular}{ |c||c|c|  }
 \hline
 \multicolumn{1}{|c||}{Networks} & \multicolumn{2}{c|}{CCC}  \\
 \hline
   & Valence & Arousal  \\
 \hline
VGGFACE1-GRU &0.52  (0.55) &0.41  (0.45)  \\
 \hline
ResNet2-GRU &0.50  (0.53) &0.42  (0.44)  \\
 \hline
DenseNet-GRU &0.48  (0.51) &0.39  (0.43)   \\
\hline
\textbf{VGGFACE1-GRU-attention} &\textbf{0.55  (0.58)} &\textbf{0.45  (0.48)}    \\
 \hline
ResNet2-GRU-attention &0.53  (0.57) &0.43  (0.45)      \\
 \hline
 DenseNet-GRU-attention &0.52  (0.57) &0.41  (0.44)   \\
\hline
\end{tabular}
\end{table}

\subsection{Experiments on RECOLA}

In the following, we evaluate the ability of the deep neural architecture developed above, trained on the Aff-Wild2, to adapt and provide highly accurate estimation of valence and arousal in the RECOLA database.  

For this reason, we fine-tuned the best performing network on Aff-Wild2, namely the VGGFACE1-GRU-attention, on the RECOLA and compared its performance with the one of two other deep neural networks that have achieved state-of-the-art performance in this database. 

The first network was an architecture comprised of a ResNet-50 and a 2-layer GRU stacked on top (let us call it ResNet-50-GRU network) as stated in \cite{1804.10938}. The second was the best performing network in the original Aff-Wild database, namely AffWildNet, as stated in \cite{kollias2,kollias3}. 

Table \ref{recola} shows the obtained results.
It is clear that the performance - for both arousal and valence estimation - of the fine-tuned VGGFACE1-GRU-attention architecture is higher than the performance of the other two networks. 
It should be mentioned that no on-the-fly or off-the-fly data augmentation \cite{kollias8,kollias9}, or post-processing techniques, were conducted when training these networks.

\begin{table}[!h]
\caption{Obtained CCC values for valence and arousal for the fine-tuned on RECOLA, best performing on Aff-Wild2, VGGFACE1-GRU-attention and for the fine-tuned on RECOLA, AffWildNet and the ResNet-GRU trained on the RECOLA.}
\label{recola}
\centering
\begin{tabular}{ |c||c|c| }
 \hline
 \multicolumn{1}{|c||}{} & \multicolumn{2}{c|}{CCC} \\
 \hline
     & Valence & Arousal  \\
 \hline
 \textbf{Fine-tuned VGGFACE1-GRU-attention} & \textbf{0.547} & \textbf{0.304}  \\
 \hline
  Fine-tuned AffWildNet \cite{kollias2,kollias3} & 0.526 & 0.273  \\
\hline
 ResNet-GRU \cite{1804.10938} & 0.462 & 0.209  \\
\hline
\end{tabular}
\end{table}

\section{CONCLUSIONS AND FUTURE WORK}\label{five}

The creation of a significant extension of the recently developed Aff-Wild database, that manages to almost double both the included videos and video frames, as well as the involved subjects, is presented in this paper. Various deep neural architectures, of the CNN-RNN type have been developed and tested on the  Aff-Wild2 database. Best performance, in valence and arousal estimation, has been obtained using a VGGFace-GRU-attention architecture, which is a CNN-RNN network followed by an attention layer. 

This architecture, that was trained with a very rich database, when fine-tuned on other existing emotion databases, such as RECOLA, has been shown to outperform  state-of-the-art networks trained on this database. 

Our future work includes extending the examined emotion analysis framework, by including different data modalities, such as audio and text, different emotion representations, such as action units and complex emotion categories, and ensembles of deep neural networks that are capable of dealing with the different types of data and emotion representations.

\bibliographystyle{ieee}
\bibliography{egbib}

\begin{thebibliography}{10}\itemsep=-1pt

\bibitem{avrithis2000broadcast}
Y.~Avrithis, N.~Tsapatsoulis, and S.~Kollias.
\newblock Broadcast news parsing using visual cues: A robust face detection
  approach.
\newblock In {\em 2000 IEEE International Conference on Multimedia and Expo.
  ICME2000. Proceedings. Latest Advances in the Fast Changing World of
  Multimedia (Cat. No. 00TH8532)}, volume~3, pages 1469--1472. IEEE, 2000.

\bibitem{cao2018vggface2}
Q.~Cao, L.~Shen, W.~Xie, O.~M. Parkhi, and A.~Zisserman.
\newblock Vggface2: A dataset for recognising faces across pose and age.
\newblock In {\em Automatic Face \& Gesture Recognition (FG 2018), 2018 13th
  IEEE International Conference on}, pages 67--74. IEEE, 2018.

\bibitem{cowie2003describing}
R.~Cowie and R.~R. Cornelius.
\newblock Describing the emotional states that are expressed in speech.
\newblock {\em Speech communication}, 40(1):5--32, 2003.

\bibitem{cowie2000feeltrace}
R.~Cowie, E.~Douglas-Cowie, S.~Savvidou*, E.~McMahon, M.~Sawey, and
  M.~Schr{\"o}der.
\newblock 'feeltrace': An instrument for recording perceived emotion in real
  time.
\newblock In {\em ISCA tutorial and research workshop (ITRW) on speech and
  emotion}, 2000.

\bibitem{cowie2012tracing}
R.~Cowie, G.~McKeown, and E.~Douglas-Cowie.
\newblock Tracing emotion: an overview.
\newblock {\em International Journal of Synthetic Emotions (IJSE)}, 3(1):1--17,
  2012.

\bibitem{dalgleish2000handbook}
T.~Dalgleish and M.~Power.
\newblock {\em Handbook of cognition and emotion}.
\newblock John Wiley \& Sons, 2000.

\bibitem{douglas2008sensitive}
E.~Douglas-Cowie, R.~Cowie, C.~Cox, N.~Amier, and D.~K. Heylen.
\newblock The sensitive artificial listner: an induction technique for
  generating emotionally coloured conversation.
\newblock In {\em LREC Workshop on Corpora for Research on Emotion and Affect}.
  ELRA, 2008.

\bibitem{ekman2002facial}
P.~Ekman.
\newblock Facial action coding system (facs).
\newblock {\em A human face}, 2002.

\bibitem{gross2010multi}
R.~Gross, I.~Matthews, J.~Cohn, T.~Kanade, and S.~Baker.
\newblock Multi-pie.
\newblock {\em Image and Vision Computing}, 28(5):807--813, 2010.

\bibitem{he2016deep}
K.~He, X.~Zhang, S.~Ren, and J.~Sun.
\newblock Deep residual learning for image recognition.
\newblock In {\em Proceedings of the IEEE Conference on Computer Vision and
  Pattern Recognition}, pages 770--778, 2016.

\bibitem{iandola2014densenet}
F.~Iandola, M.~Moskewicz, S.~Karayev, R.~Girshick, T.~Darrell, and K.~Keutzer.
\newblock Densenet: Implementing efficient convnet descriptor pyramids.
\newblock {\em arXiv preprint arXiv:1404.1869}, 2014.

\bibitem{jung2015joint}
H.~Jung, S.~Lee, J.~Yim, S.~Park, and J.~Kim.
\newblock Joint fine-tuning in deep neural networks for facial expression
  recognition.
\newblock In {\em Proceedings of the IEEE International Conference on Computer
  Vision}, pages 2983--2991, 2015.

\bibitem{koelstra2012deap}
S.~Koelstra, C.~Muhl, M.~Soleymani, J.-S. Lee, A.~Yazdani, T.~Ebrahimi, T.~Pun,
  A.~Nijholt, and I.~Patras.
\newblock Deap: A database for emotion analysis; using physiological signals.
\newblock {\em IEEE Transactions on Affective Computing}, 3(1):18--31, 2012.

\bibitem{kollias8}
D.~Kollias, S.~Cheng, M.~Pantic, and S.~Zafeiriou.
\newblock Photorealistic facial synthesis in the dimensional affect space.
\newblock In {\em Proceedings of the European Conference on Computer Vision
  (ECCV)}, pages 0--0, 2018.

\bibitem{kollias9}
D.~Kollias, S.~Cheng, E.~Ververas, I.~Kotsia, and S.~Zafeiriou.
\newblock Generating faces for affect analysis.
\newblock {\em arXiv preprint arXiv:1811.05027}, 2018.

\bibitem{kollias10}
D.~Kollias, G.~Marandianos, A.~Raouzaiou, and A.-G. Stafylopatis.
\newblock Interweaving deep learning and semantic techniques for emotion
  analysis in human-machine interaction.
\newblock In {\em 2015 10th International Workshop on Semantic and Social Media
  Adaptation and Personalization (SMAP)}, pages 1--6. IEEE, 2015.

\bibitem{kollias2}
D.~Kollias, M.~A. Nicolaou, I.~Kotsia, G.~Zhao, and S.~Zafeiriou.
\newblock Recognition of affect in the wild using deep neural networks.
\newblock In {\em Proceedings of the IEEE Conference on Computer Vision and
  Pattern Recognition Workshops}, pages 26--33, 2017.

\bibitem{kollias11}
D.~Kollias, A.~Tagaris, and A.~Stafylopatis.
\newblock On line emotion detection using retrainable deep neural networks.
\newblock In {\em 2016 IEEE Symposium Series on Computational Intelligence
  (SSCI)}, pages 1--8. IEEE, 2016.

\bibitem{kollias13}
D.~Kollias, A.~Tagaris, A.~Stafylopatis, S.~Kollias, and G.~Tagaris.
\newblock Deep neural architectures for prediction in healthcare.
\newblock {\em Complex \& Intelligent Systems}, 4(2):119--131, 2018.

\bibitem{1804.10938}
D.~Kollias, P.~Tzirakis, M.~A. Nicolaou, A.~Papaioannou, G.~Zhao, B.~Schuller,
  I.~Kotsia, and S.~Zafeiriou.
\newblock Deep affect prediction in-the-wild: Aff-wild database and challenge,
  deep architectures, and beyond, 2018.

\bibitem{kollias3}
D.~Kollias, P.~Tzirakis, M.~A. Nicolaou, A.~Papaioannou, G.~Zhao, B.~Schuller,
  I.~Kotsia, and S.~Zafeiriou.
\newblock Deep affect prediction in-the-wild: Aff-wild database and challenge,
  deep architectures, and beyond.
\newblock {\em International Journal of Computer Vision}, 127(6-7):907--929,
  2019.

\bibitem{kollias12}
D.~Kollias, M.~Yu, A.~Tagaris, G.~Leontidis, A.~Stafylopatis, and S.~Kollias.
\newblock Adaptation and contextualization of deep neural network models.
\newblock In {\em 2017 IEEE Symposium Series on Computational Intelligence
  (SSCI)}, pages 1--8. IEEE, 2017.

\bibitem{kollias7}
D.~Kollias and S.~Zafeiriou.
\newblock A multi-component cnn-rnn approach for dimensional emotion
  recognition in-the-wild.
\newblock {\em arXiv preprint arXiv:1805.01452}, 2018.

\bibitem{kollias5}
D.~Kollias and S.~Zafeiriou.
\newblock A multi-task learning \& generation framework: Valence-arousal,
  action units \& primary expressions.
\newblock {\em arXiv preprint arXiv:1811.07771}, 2018.

\bibitem{kollias6}
D.~Kollias and S.~Zafeiriou.
\newblock Training deep neural networks with different datasets in-the-wild:
  The emotion recognition paradigm.
\newblock In {\em 2018 International Joint Conference on Neural Networks
  (IJCNN)}, pages 1--8. IEEE, 2018.

\bibitem{kollias14}
D.~Kollias and S.~Zafeiriou.
\newblock Exploiting multi-cnn features in cnn-rnn based dimensional emotion
  recognition on the omg in-the-wild dataset.
\newblock {\em arXiv preprint arXiv:1910.01417}, 2019.

\bibitem{kollias15}
D.~Kollias and S.~Zafeiriou.
\newblock Expression, affect, action unit recognition: Aff-wild2, multi-task
  learning and arcface.
\newblock {\em arXiv preprint arXiv:1910.04855}, 2019.

\bibitem{lee2002welcome}
A.~Lee.
\newblock Welcome to virtualdub. org!-virtualdub. org, 2002.

\bibitem{lucey2010extended}
P.~Lucey, J.~F. Cohn, T.~Kanade, J.~Saragih, Z.~Ambadar, and I.~Matthews.
\newblock The extended cohn-kanade dataset (ck+): A complete dataset for action
  unit and emotion-specified expression.
\newblock In {\em Computer Vision and Pattern Recognition Workshops (CVPRW),
  2010 IEEE Computer Society Conference on}, pages 94--101. IEEE, 2010.

\bibitem{marsella2014computationally}
S.~Marsella and J.~Gratch.
\newblock Computationally modeling human emotion.
\newblock {\em Communications of the ACM}, 57(12):56--67, 2014.

\bibitem{mathias2014face}
M.~Mathias, R.~Benenson, M.~Pedersoli, and L.~Van~Gool.
\newblock Face detection without bells and whistles.
\newblock In {\em European Conference on Computer Vision}, pages 720--735.
  Springer, 2014.

\bibitem{mckeown2012semaine}
G.~McKeown, M.~Valstar, R.~Cowie, M.~Pantic, and M.~Schr{\"o}der.
\newblock The semaine database: Annotated multimodal records of emotionally
  colored conversations between a person and a limited agent.
\newblock {\em Affective Computing, IEEE Transactions on}, 3(1):5--17, 2012.

\bibitem{pantic2005web}
M.~Pantic, M.~Valstar, R.~Rademaker, and L.~Maat.
\newblock Web-based database for facial expression analysis.
\newblock In {\em Multimedia and Expo, 2005. ICME 2005. IEEE International
  Conference on}, pages 5--pp. IEEE, 2005.

\bibitem{parkhi2015deep}
O.~M. Parkhi, A.~Vedaldi, and A.~Zisserman.
\newblock Deep face recognition.
\newblock In {\em BMVC}, volume~1, page~6, 2015.

\bibitem{plutchik1980emotion}
R.~Plutchik.
\newblock {\em Emotion: A psychoevolutionary synthesis}.
\newblock Harpercollins College Division, 1980.

\bibitem{ringeval2017avec}
F.~Ringeval, B.~Schuller, M.~Valstar, J.~Gratch, R.~Cowie, S.~Scherer,
  S.~Mozgai, N.~Cummins, M.~Schmi, and M.~Pantic.
\newblock Avec 2017--real-life depression, and a ect recognition workshop and
  challenge.
\newblock 2017.

\bibitem{ringeval2013introducing}
F.~Ringeval, A.~Sonderegger, J.~Sauer, and D.~Lalanne.
\newblock Introducing the recola multimodal corpus of remote collaborative and
  affective interactions.
\newblock In {\em Automatic Face and Gesture Recognition (FG), 2013 10th IEEE
  International Conference and Workshops on}, pages 1--8. IEEE, 2013.

\bibitem{russell1978evidence}
J.~A. Russell.
\newblock Evidence of convergent validity on the dimensions of affect.
\newblock {\em Journal of personality and social psychology}, 36(10):1152,
  1978.

\bibitem{sneddon2012belfast}
I.~Sneddon, M.~McRorie, G.~McKeown, and J.~Hanratty.
\newblock The belfast induced natural emotion database.
\newblock {\em IEEE Transactions on Affective Computing}, 3(1):32--41, 2012.

\bibitem{soleymani2012multimodal}
M.~Soleymani, J.~Lichtenauer, T.~Pun, and M.~Pantic.
\newblock A multimodal database for affect recognition and implicit tagging.
\newblock {\em IEEE Transactions on Affective Computing}, 3(1):42--55, 2012.

\bibitem{tagaris1}
A.~Tagaris, D.~Kollias, and A.~Stafylopatis.
\newblock Assessment of parkinson’s disease based on deep neural networks.
\newblock In {\em International Conference on Engineering Applications of
  Neural Networks}, pages 391--403. Springer, 2017.

\bibitem{tagaris2}
A.~Tagaris, D.~Kollias, A.~Stafylopatis, G.~Tagaris, and S.~Kollias.
\newblock Machine learning for neurodegenerative disorder diagnosis—survey of
  practices and launch of benchmark dataset.
\newblock {\em International Journal on Artificial Intelligence Tools},
  27(03):1850011, 2018.

\bibitem{tian2001recognizing}
Y.-l. Tian, T.~Kanade, and J.~F. Cohn.
\newblock Recognizing action units for facial expression analysis.
\newblock {\em Pattern Analysis and Machine Intelligence, IEEE Transactions
  on}, 23(2):97--115, 2001.

\bibitem{valstar2010induced}
M.~Valstar and M.~Pantic.
\newblock Induced disgust, happiness and surprise: an addition to the mmi
  facial expression database.
\newblock In {\em Proc. 3rd Intern. Workshop on EMOTION (satellite of LREC):
  Corpora for Research on Emotion and Affect}, page~65, 2010.

\bibitem{whissel1989dictionary}
C.~Whissel.
\newblock The dictionary of affect in language, emotion: Theory, research and
  experience: vol. 4, the measurement of emotions, r.
\newblock {\em Plutchik and H. Kellerman, Eds., New York: Academic}, 1989.

\bibitem{yin2008high}
L.~Yin, X.~Chen, Y.~Sun, T.~Worm, and M.~Reale.
\newblock A high-resolution 3d dynamic facial expression database.
\newblock In {\em Automatic Face \& Gesture Recognition, 2008. FG'08. 8th IEEE
  International Conference On}, pages 1--6. IEEE, 2008.

\bibitem{yin20063d}
L.~Yin, X.~Wei, Y.~Sun, J.~Wang, and M.~J. Rosato.
\newblock A 3d facial expression database for facial behavior research.
\newblock In {\em Automatic face and gesture recognition, 2006. FGR 2006. 7th
  international conference on}, pages 211--216. IEEE, 2006.

\bibitem{zafeiriou2017aff}
S.~Zafeiriou, D.~Kollias, M.~A. Nicolaou, A.~Papaioannou, G.~Zhao, and
  I.~Kotsia.
\newblock Aff-wild: Valence and arousal ‘in-the-wild’challenge.
\newblock In {\em 2017 IEEE Conference on Computer Vision and Pattern
  Recognition Workshops (CVPRW)}, pages 1980--1987. IEEE, 2017.

\bibitem{kollias1}
S.~Zafeiriou, D.~Kollias, M.~A. Nicolaou, A.~Papaioannou, G.~Zhao, and
  I.~Kotsia.
\newblock Aff-wild: Valence and arousal'in-the-wild'challenge.
\newblock In {\em Proceedings of the IEEE Conference on Computer Vision and
  Pattern Recognition Workshops}, pages 34--41, 2017.

\bibitem{zhou2015lie}
Y.~Zhou, H.~Zhao, and X.~Pan.
\newblock Lie detection from speech analysis based on k--svd deep belief
  network model.
\newblock In {\em International Conference on Intelligent Computing}, pages
  189--196. Springer, 2015.

\end{thebibliography}

\end{document}